\documentclass [a4paper, 11pt, twocolumn, twoside]{article}

\usepackage{amsmath, amsfonts, amssymb, latexsym, bm}
\usepackage{cancel}

\usepackage{color}
\usepackage[dvipsnames]{xcolor}
\usepackage{graphics, graphicx}
\usepackage{tikz} 

\usepackage{fancyhdr}
\usepackage{multicol, multirow}
\usepackage{makecell}
\usepackage{array,booktabs}
\usepackage{mhchem}

\usepackage{listings}
\usepackage{algorithm}
\usepackage{algpseudocode}

\usepackage[hidelinks]{hyperref}
\usepackage{breakurl}
\usepackage{authblk}

\usepackage[textsize=tiny, textwidth=2.4cm]{todonotes}

\usepackage{ngerman}    
\newcommand{\ConfName}{\textit{Scientific Computing 2023}}

\title{Bringing Chemistry to Scale: Loss Weight Adjustment for Multivariate Regression in Deep Learning of Thermochemical Processes}
\author[1]{Franz M. Rohrhofer}
\author[2]{Stefan Posch}
\author[2]{Clemens Gößnitzer}
\author[3]{José M. García-Oliver}
\author[1]{Bernhard C. Geiger}
\affil[1]{\small Know-Center GmbH -- Research Center for Data-Driven Business \& Big Data Analytics, Sandgasse 36/4, 8010 Graz, Austria}
\affil[2]{\small LEC GmbH -- Large Engines Competence Center, Inffeldgasse 19, 8010 Graz, Austria}
\affil[3]{\small CMT Motores Térmicos - Universitat Politècnica de València, Camí de Vera, 46022 València, Spain}
\date{} 

\newcommand{\authorshort}{Franz M. Rohrhofer et al.}
\newcommand{\titleshort}{Loss Weight Adjustment for Multivariate Regression in DL}

\begin{document}

\maketitle\thispagestyle{plain}

\begin{abstract}
Flamelet models are widely used in computational fluid dynamics to simulate thermochemical processes in turbulent combustion. 
These models typically employ memory-expensive lookup tables that are predetermined and represent the combustion process to be simulated.
Artificial neural networks (ANNs) offer a deep learning approach that can store this tabular data using a small number of network weights, potentially reducing the memory demands of complex simulations by orders of magnitude.
However, ANNs with standard training losses often struggle with underrepresented targets in multivariate regression tasks, e.g., when learning minor species mass fractions as part of lookup tables.
This paper seeks to improve the accuracy of an ANN when learning multiple species mass fractions of a hydrogen (\ce{H2}) combustion lookup table. 
We assess a simple, yet effective loss weight adjustment that outperforms the standard mean-squared error optimization and enables accurate learning of all species mass fractions, even of minor species where the standard optimization completely fails. 
Furthermore, we find that the loss weight adjustment leads to more balanced gradients in the network training, which explains its effectiveness.
\end{abstract}	

\section{Introduction} 
\label{sec:intro}
Driven by the urgent need for more efficient and less polluting combustion engines, understanding and modeling of thermochemical processes through computational fluid dynamics (CFD) has become an essential part in engineering. 
To cope with the huge computational demands of including detailed chemical reaction mechanisms in three-dimensional simulations, several approaches for simulating turbulent combustion have been developed over the past decades. 
Flamelet models are widely used in CFD and belong to a category of approaches which may utilize indexing of a predetermined lookup table to accelerate calculations during simulation run-time.
These lookup tables contain all the necessary information of the thermochemical process and can take up several gigabytes of memory space, potentially hindering their applicability for complex simulations when resources are scarce.

With the recent upsurge of scientific deep learning (DL), the use of ANNs has become popular, since ANNs can store the information of lookup tables with a finite, possibly small, number of network weights.
Several approaches have been introduced, and differ in the choice of input (control) variables and predicted (thermochemical) scalars of the lookup table.
Recently, \cite{mousemi2023application} has successfully demonstrated the use of an ANN-based approach in the simulation of a turbulent premixed flame, while reducing the memory size of the flamelet table by around $92\%$.
Furthermore, \cite{honzawa2021experimental} also employed an ANN model for simulating water-sprayed turbulent combustion and reported reduced memory demands, while achieving a comparable accuracy to that of a conventional flamelet-generated manifold (FGM) approach.

However, a common challenge in DL for tabulating lookup tables is simultaneously learning and predicting multiple species mass fractions with a single ANN. 
ANNs performing a multivariate regression task often struggle to accurately predicting minor species, whose target mass fractions are much smaller than those of major species.
This issue was explicitly reported in~\cite{owoyele2020application} and~\cite{ding2022machine}, and is usually addressed by learning smaller groups of species with similar orders of magnitude of mass fraction.
Such clustering approach is followed, e.g., in~\cite{ding2021machine} by the multiple multilayer perceptron (MMP) approach, in~\cite{owoyele2021efficient} by the mixture of experts (MoE) approach or in~\cite{ranade2021efficient} by self-organizing maps (SOM).
Overall, these methods are effective in learning species mass fractions of lookup tables, but they often come with an increased computational expense since multiple ANN instances have to be trained.

This work aims at improving the overall accuracy of a single ANN which is trained on multiple species mass fractions of thermochemical processes.
To cope with the different ranges of target values, we employ a simple loss weight adjustment and test this approach in learning a lookup table of \ce{H2} combustion which involves nine species.
This system is chosen due to the current interest for combustion applications.
We compare the accuracy of the loss weight adjustment with the use of standard mean-squared error (MSE) optimization.
We study backpropagated gradients in the optimization and observe that the loss weight adjustment leads to more balanced gradients, which explains its effectiveness in learning multiple species with different target mass fractions.

\section{Background}

\subsection{Flamelet Model}\label{sec:flamelet_model}
The flamelet model, initially derived by Peters~\cite{peters1988laminar}, assumes that a turbulent flame is an ensemble of laminar one-dimensional reacting structures, referred to as 'flamelets'.  
Flamelets are solved externally in ad-hoc solvers, from which the thermochemical state in terms of temperature $T$ and species mass fraction $Y_j$ is retrieved and later fed into the CFD code.
This approach overcomes the need to solve a transport equation for each species in every cell of the computational domain.
Some flamelet methods are tabulated to allow for accelerated CFD-flamelet interactions during simulation run-time. 
Depending on the flame configuration, different approaches may be followed; 
for example, in the current approach an Unsteady Flamelet Progress Variable (UFPV) tabulation depending on mixture fraction $Z$ (local fuel-air ratio), progress variable $C$ (chemical evolution from inert to steady conditions) and scalar dissipation rate $\chi$ (related to the spatial gradient along the 1D flamelet domain) is followed~\cite{naud2015rans}.

\subsection{Artificial Neural Networks}
ANNs are the 'work horse' of deep learning, due to their ability of approximating any continuous function given sufficient expressive power (universal approximation theorem).
Their flexibility has rendered ANNs suitable for substituting lookup tables, i.e., to learn the nonlinear mapping $\bm{X}\mapsto\bm{Y}$, with $\bm{X}\in\mathbb{R}^D$ denoting the input control variables and $\bm{Y}\in\mathbb{R}^d$ the approximated thermochemical scalars, in this work the species mass fractions $Y_j$.

\subsubsection{Deep Multilayer Perceptron}
Multilayer perceptrons (MLPs) are a simple class of feedforward ANNs with multiple layers of perceptrons (or neurons) which are fully-connected.
For a fully-connected MLP with $L$ layers, the network's output is given by the recursive application of activations
\begin{equation}
\bm{Y}=f_L(\bm{W}_L,\dotsm, f_2(\bm{W_2}, f_1(\bm{W}_1, \bm{X}))\dotsm),
\end{equation}
where $\bm{W}_l$ contains the weights and biases $\theta_{i,l}$ of layer $l$, and $f_l$ denotes the neuron's activation function of that layer.
The activation functions are (except for the final layer) nonlinear where common choices for regression tasks are the hyperbolic tangent ($\tanh$) or Sigmoid linear unit (swish).
For simplicity, we denote the weights and biases of the full network function with a single weight vector $\bm{\theta}$.

\subsubsection{Loss Function \& Optimization}\label{sec:loss_and_optimization}
To train the ANN on a specific task, the selection of an appropriate loss function is necessary. 
A common choice for regression tasks in DL is the well-known MSE loss
\begin{equation}
    \mathcal{L}(\bm{\theta})=\frac{1}{N} \sum_{i=1}^N \left( \bm{Y}^{(i)} - \bm{Y}_{\bm{\theta}}^{(i)} \right)^2, \label{eq:MSE}
\end{equation}
with training examples $\left\{ (\bm{X}^{(i)}, \bm{Y}^{(i)}) \right\}_{i=1}^N$ and $\bm{Y}_{\bm{\theta}}$ denoting the network's output.
The optimization, i.e., finding the network weights that minimize the loss function, is performed in an iterative (gradient-based) process and according to the general update rule
\begin{equation}
    \bm{\theta}_{n+1} \leftarrow \bm{\theta}_n - \alpha\frac{\partial \mathcal{L}}{\partial \bm{\theta}}, \label{eq:update_rule}
\end{equation}
where $\alpha$ denotes the learning rate.
The gradients of the loss function w.r.t.\ the network weights can be calculated via the backpropagation algorithm that is part of modern deep learning libraries.

\section{Methods}\label{sec:methods}
This section provides details on the loss weight adjustment that will be used in our experiments. 
We start by splitting the multivariate MSE loss from Equation~\eqref{eq:MSE} into the individual species mass fractions $\bm{Y}=(Y_0, Y_1, \dots, Y_d)^T$ 
\begin{equation}
    \mathcal{L}(\bm{\theta})=\frac{1}{N} \sum_{i=1}^N \sum_{j=1}^d \left( Y_j^{(i)} - Y_{j,\bm{\theta}}^{(i)} \right)^2, 
    \label{eq:MSE_detailed}
\end{equation}
where $Y_{j,\bm{\theta}}$ is the network's output of the $j$-th species mass fraction.
Furthermore, we can write
\begin{equation}
    \mathcal{L}(\bm{\theta}) = \sum_{j=1}^d \mathcal{L}_j(\bm{\theta}),
    \label{eq:MSE_components}
\end{equation}
where $\mathcal{L}_j$ is the univariate MSE, cf. Equation~\eqref{eq:MSE}, for each individual species mass fraction $Y_j$.
From this perspective, the use of the MSE optimization in multivariate regression can be seen as a single objective by (linear) scalarization of the multiple objectives, where each objective is given by the optimization of a single target species mass fraction with a weight equal to one.
Equations~\eqref{eq:MSE},~\eqref{eq:MSE_detailed} and~\eqref{eq:MSE_components} denote the same loss function which we will refer to as \textit{standard} MSE for the later course of this work.
According to Equation~\eqref{eq:MSE_components}, the gradient update rule can be written in the form
\begin{equation}
    \bm{\theta}_{n+1} \leftarrow \bm{\theta}_n - \alpha \sum_{j=1}^d \frac{\partial \mathcal{L}_j}{\partial \bm{\theta}}, \label{eq:update_rule_MO}
\end{equation}
which implies that different orders of loss magnitudes yield different contributions to the gradient update.
This can have serious consequences for the final training outcome as it will be demonstrated in the later part of this work.

\subsection{Loss Weight Adjustment}\label{sec:loss_weight_adjustment}
In general, loss weights are commonly used in DL, ranging from Lagrange multipliers that adjust the strength of $L_1$ (lasso) or $L_2$ (ridge) regularization, to hyperparameters that adjust the importance of different objectives in physics-informed neural networks~\cite{rohrhofer2021pareto} or variational autoencoders~\cite{kingma2019introduction}.
For our porpose, we note that the MSE loss is bound to absolute ranges of targets.
Hence, by assuming that the ANN induces equal relative errors across the different species mass fractions, the standard MSE and respective gradients are potentially dominated by species with larger mass fractions.
To compensate this, we consider weighting the individual components in Equation~\eqref{eq:MSE_components} according to 
\begin{equation}
    \mathcal{L}(\bm{\theta}) = \sum_{j=1}^d \omega_j \mathcal{L}_j(\bm{\theta}), \label{eq:weighted_MSE}
\end{equation}
where $\omega_j$ is a scalar weight.
Thus, individual gradients in the update rule are also scaled by 
\begin{equation}
    \bm{\theta}_{n+1} \leftarrow \bm{\theta}_n - \alpha \sum_{j=1}^d \omega_j\frac{\partial \mathcal{L}_j}{\partial \bm{\theta}}. \label{eq:weighted_update_rule_MO}
\end{equation}
The choice of appropriate loss weights $\omega_j$ is crucial for the success of this approach.
In this work, we choose the variance as a measure of dispersion, to determine how far individual mass fractions $Y_j$ are spread out from their average value $\bar Y_j$
\begin{equation}
    \omega_j := \frac{1}{\mathrm{Var}(Y_j)},
\end{equation}
with $\mathrm{Var}(Y_j)=\sum_{i=1}^N(Y_j^{(i)}-\bar Y_j)^2/N$.
This effectively assigns greater loss weights to minor species with smaller mass fractions compared to those with larger mass fractions.
Furthermore, determining the variance of species mass fractions has to be performed only once prior network training which is a computationally cheap and robust method for selecting adequate loss weights. 
The weights can be applied either directly at the loss computation in Equation~\eqref{eq:weighted_MSE} or in the gradient update rule in Equation~\eqref{eq:weighted_update_rule_MO} using the standard MSE.
Both methods effectively scale the gradients in the final optimization step and will be later on referred to as \textit{weighted} MSE optimization.

\section{Experimental Setting}\label{sec:experimental_setup}

\subsection{Datasets}
For our experiments we use a single databases, representing the lookup table for \ce{H2} combustion, as computed with the flamelet method discussed in Section~\ref{sec:flamelet_model}.
The database contains 6.7~million data points with nine species mass fractions.
Details on the involved species can be taken from Table~\ref{tab:H2}.
We choose as input the progress variable $C$, the mixture fraction $Z$ and scalar dissipation rate $\chi$ and approximate $\bm{Y}(C,Z,\chi)$ with a single ANN where $\bm{Y}\in\mathbb{R}^9$.
A 80/10/10 dataset split is used where $80\%$ of the database is taken for training, and $10\%$ for validation and testing, respectively.

\begin{figure*}[t]
\begin{center}
\centerline{\includegraphics[width=\textwidth]{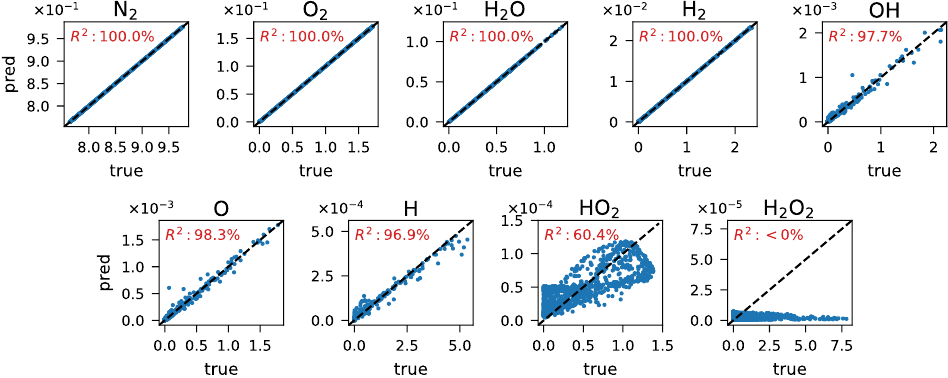}}
\caption{\textbf{Standard MSE Optimization.} True vs. predicted species mass fractions for the nine species in the $\mathrm{H}_2$ test set. A perfect fit is indicated by the diagonal, dashed line. The accuracy in terms of the $R^2$ score is shown in red. Minor species, such as \ce{HO2} or \ce{H2O2}, are not learned accurately.}    
\label{fig:H2_none_pred}
\end{center}
\end{figure*}
\subsection{Network Architecture \& Optimization}
We use fully-connected ANNs with four hidden layers, 50 neurons per layer and $\tanh$ activation function.
For the output neurons, we use a softmax logistic layer to account for mass conservation in the species mass fractions, since $\sum_j Y_j = 1$.
The network weights are initialized using Glorot initialization and inputs are scaled using MinMax feature scaling.
The ANNs are optimized with stochastic gradient-descent on mini-batches of size 1024, i.e., the loss function for the standard or weighted optimization is computed over 1024 consecutive training points.
We choose a learning rate of $\alpha=0.001$ and perform 50 iterations (epochs) through the entire training dataset.

\section{Results}
Our experiments are arranged as follows:
first, in Section~\ref{sec:standard_MSE} we train and test a single ANN on the \ce{H2} database by using the standard MSE optimization as discussed in Section~\ref{sec:loss_and_optimization}.
Next, the loss weight adjustment, as introduced in Section~\ref{sec:loss_weight_adjustment}, is applied and results are presented in Section~\ref{sec:weighted_MSE}. 
Finally, in Section~\ref{sec:summary} we extend and compare the two optimization approaches on different splits of training and validation set.
We measure the final accuracy in terms of coefficient of determination 
\begin{equation}
    R^2=1-\frac{\sum_i^N\left(Y_j^{(i)}-Y_{j,\bm{\theta}}^{(i)}\right)^2}{\sum_i^N\left(Y_j^{(i)}-\bar Y_j\right)^2},
\end{equation}
for each species mass fraction $Y_j$ on the test set, and use the validation set to record the MSE of each individual species and backpropagated gradients during optimization.

\begin{figure}[t]
\begin{center}
\centerline{\includegraphics[width=\columnwidth]{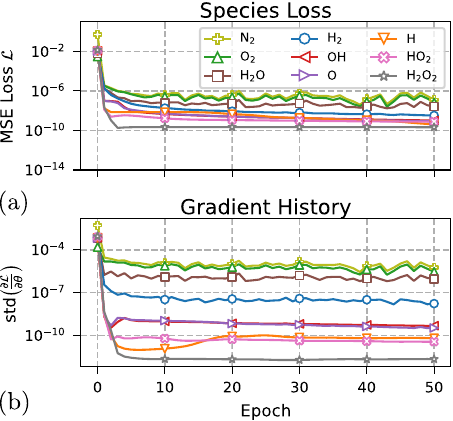}}
\caption{\textbf{Standard MSE Optimization.} Recorded (a) species loss and (b) backpropagated gradients for each of the nine species in the \ce{H2} database. The unbalanced gradients during optimization explain the issue in learning the minor species, in particular \ce{H2O2} (cf Figure~\ref{fig:H2_none_pred}).}    
\label{fig:H2_none_hist}
\end{center}
\end{figure}

\begin{figure}[t]
\begin{center}
\centerline{\includegraphics[width=\columnwidth]{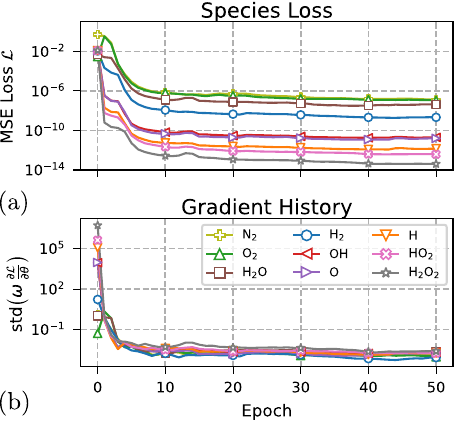}}
\caption{\textbf{Weighted MSE Optimization.} Recorded (a) species loss and (b) backpropagated gradients for each of the nine species in the \ce{H2} database. The balanced gradients during optimization explain the success in a equal and continuous learning of all species (cf Figure~\ref{fig:H2_var_pred}).}    
\label{fig:H2_var_hist}
\end{center}
\end{figure}

\subsection{Standard MSE Optimization}\label{sec:standard_MSE}
For the first experiment, we train a single ANN on the \ce{H2} database using the standard MSE optimization.
Figure~\ref{fig:H2_none_pred} shows the true versus predicted species mass fractions of the nine species in the \ce{H2} database, evaluated on the test set.
A perfect fit is indicated by the diagonal, dashed line and the accuracy (in terms of $R^2$) is given in the upper left corner.
We note that the order of species is based on absolute ranges of species mass fractions, decreasing from left to right and top to bottom. 
From the figure it is evident that, while major species such as \ce{N2}, \ce{O2}, \ce{H2O} and \ce{H2} are predicted well ($R^2>99\%$), the accuracy drops considerably for minor species, in particular for \ce{H2O2} with a $R^2$ value below zero.
To investigate this issue further, we refer to Figure~\ref{fig:H2_none_hist} which shows the individual losses and backpropagated gradients, evaluated on the validation set during the optimization.  
For simplicity, we use the standard deviation of backpropagated gradients as a general measure of gradient information that updates the network weights.
In Figure~\ref{fig:H2_none_hist}b, we observe unbalanced gradients across the different species, with minor species having significant lower gradients than major species.
The discrepancy originates from the spread in the individual losses where, as already discussed in Section~\ref{sec:methods}, the contribution from major species to the loss and gradient update is significantly larger.
This explains why mass fractions of major species in Figure~\ref{fig:H2_none_pred} are learned more accurately than that of minor species, even when the network size or learning rate is varied (also tested but not shown).

\begin{figure*}[t]
\begin{center}
\centerline{\includegraphics[width=\textwidth]{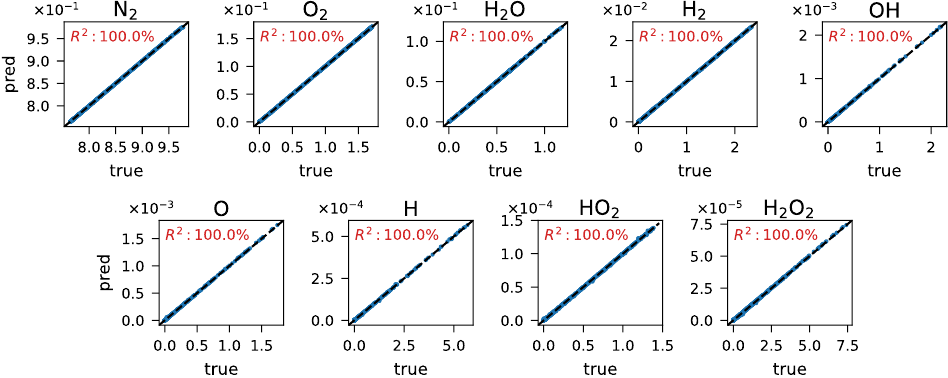}}
\caption{\textbf{Weighted MSE Optimization.} True vs. predicted species mass fractions for the nine species in the \ce{H2} test set. A perfect fit is indicated by the diagonal, dashed line. The accuracy in terms of the $R^2$ score is shown in red. All species in the database are learned perfectly ($R^2\approx100$), even minor ones.}    
\label{fig:H2_var_pred}
\end{center}
\end{figure*}
\subsection{Weighted MSE Optimization}\label{sec:weighted_MSE}
We now perform the same training and testing procedure as before, but this time apply the loss weight adjustment explained in Section~\ref{sec:loss_weight_adjustment}.
Figure~\ref{fig:H2_var_pred} shows the true versus predicted species mass fractions of the nine species in the \ce{H2} database, evaluated on the test set.
We observe a nearly perfect fit ($R^2\approx100\%$) of all species mass fractions, even for minor species.
Figure~\ref{fig:H2_var_hist} shows the individual losses and backpropagated gradients, evaluated on the validation set during the optimization.  
By looking at the individual losses (Figure~\ref{fig:H2_var_hist}a), it is evident that, in comparison to the previous results, there is an equal and continuous improvement across all species.
This can be explained by Figure~\ref{fig:H2_var_hist}b which shows that backpropagated gradients are mostly in balance during optimization and, hence, account for an equal and continuous learning of all species mass fractions. 
We note that although gradients start at a much different range when the loss weight adjustment is applied, the same initial learning rate of $\alpha=0.001$ still works well as a decent step size that results in balanced gradients during optimization.

\begin{table*}[t]
\caption{ANN performance ($R^2$, mean and standard deviation over ten uniquely trained instances) on the nine species of the \ce{H2} database. $R^2$ scores below zero are listed as $<0$. Significant improvement on minor species is achieved when the loss weight adjustment is applied.}
\label{tab:H2}
\newcolumntype{C}{ @{}>{${}}c<{{}$}@{} }
\renewcommand{\arraystretch}{1.2}
\newcommand{\mc}[1]{\multicolumn{3}{c}{#1}}
\resizebox{\textwidth}{!}{
\begin{tabular}{l *9{rCl} }
    \Xhline{3\arrayrulewidth}
    \textbf{Specie} & \mc{\textbf{\ce{N2}}} & \mc{\textbf{\ce{O2}}} & \mc{\textbf{\ce{H2O}}} & \mc{\textbf{\ce{H2}}} & \mc{\textbf{\ce{OH}}} & \mc{\textbf{\ce{O}}} & \mc{\textbf{\ce{H}}} & \mc{\textbf{\ce{HO2}}} & \mc{\textbf{\ce{H2O2}}}
    \\ 
    \bm{$\mathrm{Var}(Y_j)$} & 6.22&\cdot&$10^{-2}$ & 5.48&\cdot& $10^{-2}$ & 2.18&\cdot&$10^{-2}$ & 6.73&\cdot&$10^{-3}$ & 2.97&\cdot&$10^{-4}$ & 2.87&\cdot&$10^{-4}$ & 8.07&\cdot&$10^{-5}$ & 4.33&\cdot&$10^{-5}$ & 1.24&\cdot&$10^{-5}$              
    \\ \hline \hline
    \textbf{standard} & 100.00&\pm&0.01 & 100.00&\pm&0.01 & 99.99&\pm&0.01 & 99.99&\pm&0.01 & 98.85&\pm&0.30 & 99.00&\pm&0.29 & 93.59&\pm&5.69 & 58.42&\pm&2.98 & \mc{$<0$}             
    \\
    \textbf{weighted} & 100.00&\pm&0.01 & 100.00&\pm&0.01 & 99.99&\pm&0.01 & 100.00&\pm&0.01 & 99.98&\pm&0.02 & 99.98&\pm&0.02 & 99.98&\pm&0.01 & 99.98&\pm&0.01 & 99.97&\pm&0.01    
    \\
    \Xhline{3\arrayrulewidth}
\end{tabular}
}
\end{table*}
\subsection{Comparison \& Summary}\label{sec:summary}
While the previous results have demonstrated the effectiveness of the loss weight adjustment on a single experimental setup, it remains unclear whether this can be extended to the general case, especially on different dataset splits.
To this end, we train and test multiple networks.
We record the average accuracy over ten different ANN instances, each uses a unique seed for the initialization of network weights, and split of training and validation set.
We optimize them with and without the loss weight adjustment.

The results can be found in Table~\ref{tab:H2}.
In the table header, each species and the respective variance of its mass fraction $\mathrm{Var}(Y_j)$ is listed.
The main table shows the mean and standard deviation of the $R^2$ score over the ten individual runs as determined on the test set.
$R^2$ scores below zero are listed as $<0$.
From the table it is evident that the use of the loss weight adjustment vastly outperforms the standard optimization in accurately learning all species, in particular minor species mass fractions where the standard optimization completely fails.
In summary, these results demonstrate the effectiveness of the loss weight adjustment which resolves issues in learning species mass fractions on different scales.

\section{Conclusion}
The use of ANNs for replacing lookup tables of flamelet models has become a prominent tool to cope with the huge computational expenses of turbulent combustion modeling. 
Training ANNs on such tabular data usually involves learning multiple species mass fractions with different characteristics. 
This work discussed a simple, yet effective loss weight adjustment that balances the loss gradients of different species and brings the ANN optimization to scale.
This significantly improved the overall accuracy of a single ANN, performing a multivariate regression task on several species with different mass fractions.
Future work will be devoted to extending this approach to more complex systems.
In general, our results provide valuable insights into using ANNs for multivariate regression tasks and may find further use in several disciplines beyond thermochemical processes, such as multiscale and multiphysics modelling.

\section*{Acknowledgments}

The authors acknowledge the financial support of the Austrian COMET — Competence Centers for Excellent Technologies — Programme of the Austrian Federal Ministry for Climate Action, Environment, Energy, Mobility, Innovation and Technology, the Austrian Federal Ministry for Digital and Economic Affairs, and the States of Styria, Upper Austria, Tyrol, and Vienna for the COMET Centers Know-Center and LEC EvoLET, respectively. The COMET Programme is managed by the Austrian Research Promotion Agency (FFG).
Research activity by José M García-Oliver was partially funded by ORIONe project (PDC2021-121066-C22) from the Agencia Estatal de Investigación of the Spanish Government.

\bibliography{my_literature}

\end{document}